\documentclass[12pt,journal,draftcls,letterpaper,onecolumn]{IEEEtran}

\usepackage{graphicx}
\usepackage{algorithm}
\usepackage{algorithmic}

\usepackage{amssymb}

\usepackage{cite}

\usepackage{tabularx}

\usepackage{amsmath}

\usepackage{amssymb}

\usepackage{mathrsfs}

\usepackage{amsfonts}

\usepackage{url}

\newcommand{\elemset}{\mathcal}
\newcommand{\UCRL}{\elemset{R}_L}
\newcommand{\UCRU}{\elemset{R}_U}
\newcommand{\UCRES}{\elemset{M}}
\newcommand{\UCX}{\elemset{X}}
\newcommand{\UCS}{\elemset{S}}
\newcommand{\newindex}[1]{\textit{#1}\index{#1}}
\newcommand{\lattice}{\mathcal}
\begin{document}
%% Begin your document with this command:

%\begin{frontmatter}

%% +---------------+
%% | ARTICLE TITLE |
%% +---------------+
% Title, authors and addresses

% use the thanksref command within \title, \author or \address for
% footnotes;
% use the corauthref command within \author for corresponding author
% footnotes;
% use the ead command for the email address,
% and the form \ead[url] for the home page:
% \title{Title\thanksref{label1}}
% \thanks[label1]{}
% \author{Name\corauthref{cor1}\thanksref{label2}}
% \ead{email address}
% \ead[url]{home page}
% \thanks[label2]{}
% \corauth[cor1]{}
% \address{Address\thanksref{label3}}
% \thanks[label3]{}

\title{A branch-and-bound feature selection algorithm for U-shaped
  cost functions}

%% +---------+
%% | AUTHORS |
%% +---------+
% use optional labels to link authors explicitly to addresses:
% \author[label1,label2]{}
% \address[label1]{}
% \address[label2]{}

\author{Marcelo Ris,
        Junior Barrera,
        David C. Martins Jr
\thanks{M. Ris, J. Barrera and D. C. Martins Jr are with the
Instituto de Matem\'{a}tica e Estat\'{i}stica, Universidade de S\~{a}o
        Paulo,
S\~{a}o Paulo, Brazil}}

\markboth{A branch-and-bound optimization algorithm for U-shaped cost
functions on Boolean lattices}
{A branch-and-bound optimization algorithm for U-shaped cost
functions on boolean lattices}

\maketitle

\begin{abstract}
This paper presents the formulation of a combinatorial optimization
problem with the following
characteristics: $i$.the search space is the power set of a finite set
structured as a Boolean lattice;
$ii$.the cost function forms a U-shaped curve when applied to any
lattice chain.
This formulation applies for feature selection in the context of
pattern recognition.
The known approaches for this problem are branch-and-bound algorithms
and heuristics, that explore partially the search
space. Branch-and-bound algorithms are equivalent to the full search,
while heuristics are not. This paper presents a branch-and-bound
algorithm that differs from the others known by exploring the lattice
structure and the U-shaped chain curves of the search space. The main
contribution of this paper is the architecture of this algorithm that
is based on the representation and exploration of the search space by
new lattice properties proven here.
Several experiments, with well known public data, indicate the
superiority of
the proposed method to SFFS, which is a popular heuristic that gives
good results in very short computational time. In all experiments, the
proposed method got better or equal results in similar or even smaller
computational time.
\end{abstract}

%\end{frontmatter}

%----------------------------------------
\section{Introduction}
A combinatorial optimization algorithm chooses the object of minimum
cost over a finite collection of objects,
called search space, according to a given cost function. The simplest
architecture for this algorithm, called full search,
access each object of the search space, but it does not work for huge
spaces. In this case, what is
possible is to access some objects and choose the one of minimum cost,
based on the observed measures.  Heuristics and branch-and-bound
are two families of algorithms of this kind. An heuristic algorithm
does not have formal guaranty of finding the minimum
cost object, while a branch-and-bound algorithm has mathematical
properties that guarantee to find it.

Here, it is studied a combinatorial optimization problem such that the
search space is composed
of all subsets of a finite set with $n$ points (i.e., a search space
with
$2^n$ objects),
organized as a Boolean lattice, and the cost function has a U-shape in
any chain of the search space or, equivalently,
the cost function has a U-shape in any maximal chain of the search space.

This structure is found in some applied problems such as feature
selection in pattern recognition \cite{dudahart2000chap1,jaduma:2000}
and
W-operator window design in mathematical morphology
\cite{dmartins2006}.
In these problems, a minimum subset of features, that is sufficient to
represent the
objects, should be chosen from a set of $n$ features. In W-operator
design, the features are points of a finite rectangle of $Z^2$ called window.
The U-shaped functions are formed by error estimation of the
classifiers or of the
operators designed or by some measures, as the entropy, on the corresponding
estimated join distribution. This is a well known phenomenon
in pattern recognition:
for a fixed amount of training data, the increasing number of features
considered in the classifier design induces the reduction
of the classifier error by increasing the separation between classes
until the available data becomes too small
to cover the classifier domain and the consequent increase of the estimation
error induces the increase of the classifier error.
Some known approaches for this problem are heuristics.
A relatively well succeeded heuristic algorithm is SFFS
\cite{pudil94},  which gives good results in relatively small
computational time.

There is a myriad of branch-and-bound algorithms in the literature
that are based on monotonicity of the
cost-function \cite{frank,nakariyakul,wang,yang}. For a detailed
review of
branch-and-bound algorithms, refer to \cite{somol}. If the real
distribution
of the joint probability between the patterns and their classes were
known,
larger dimensionality would imply in smaller classification
errors. However, in practice,
these distributions are unknown and should be estimated. A problem
with
the adoption of monotonic cost-functions is that they do not take into
account the estimation errors committed when many features are
considered (``curse of dimensionality'' also known as ``U-curve
problem'' or ``peaking phenomena'' \cite{jaduma:2000}).

This paper presents a branch-and-bound algorithm that differs from the
others known by exploring the
lattice structure and the
U-shaped chain curves of the search space.

Some experiments were performed to compare the SFFS to the
U-curve approach. Results obtained from applications such as
W-operator window design, genetic network architecture identification
and eight UCI repository data sets show encouraging results, since the
U-curve algorithm beats (i.e., finds a node with smaller cost than the
one found by SFFS) the SFFS results in smaller computational time for
27 out of 38 data sets tested. For all data sets, the
U-curve algorithm gives a result equal or better than SFFS, since the
first covers the complete search space.

Though the results obtained with the application of the method
developed to pattern recognition problems are exciting, the great
contribution of this paper is the discovery of some lattice algebra
properties that lead to a new data structure for the search space
representation, that is particularly adequate for updates after
up-down lattice interval cuts (i.e., cuts by couples of intervals
[0,X] and [X,W]). Classical tree based search space representations
does not have this property. For example, if the Depth First Search
were adopted to represent the Boolean lattice only cuts in one
direction could be performed.

Following this introduction, Section 2 presents the formalization of
the problem studied. Section 3 describes structurally
the branch-and-bound algorithm designed. Section 4 presents the
mathematical properties that support the algorithm steps.
Section 5 presents some experimental results comparing U-curve to
SFFS.
Finally, Conclusion discusses the contributions of this paper and
proposes some next steps of this research.

%----------------------------------------
\section{The Boolean U-curve optimization problem}
Let $W$ be a finite subset,  $\mathscr{P}(W)$ be the collection of all
subsets of $W$, $\subseteq$ be the
usual inclusion relation on sets and, $|W|$ denote the cardinality of
$W$. The search space is composed by $2^{|W|}$ objects organized in a
Boolean lattice.

The partially ordered set $(\mathscr{P}(W), \subseteq)$ is a complete
Boolean lattice of degree $|W|$ such that: the
smallest and largest elements are, respectively, $\emptyset$ and $W$;
the sum and product are, respectively, the usual union
and intersection on sets and the complement of a set $X$ in
$\mathscr{P}(W)$ is its complement in relation to $W$,
denoted by $X^c$.

Subsets of $W$ will be represented by strings of zeros and ones, with
$0$ meaning that the point does not belong to the subset
and $1$ meaning that it does. For example, if $W = \{ (-1,0),(0,0),$
$(+1,0) \}$, the subset
$\{ (-1,0),(0,0)\}$ will be represented by $110$. In an abuse of
language, $X = 110$ means that $X$ is the set represented by $110$.

A chain $\elemset{A}$ is a collection
$\{A_1, A_2, \ldots, A_k\} \subseteq \UCX \subseteq \mathscr{P}(W)$
such that
$A_1 \subseteq A_2 \subseteq \ldots \subseteq A_k$.
A chain $\elemset{M} \subseteq \UCX$ is maximal in $\UCX$ if
there is no other chain $\elemset{C} \subseteq \UCX$ such that
$\elemset{C}$ contains properly $\elemset{M}$.

Let $c$ be a cost function defined from $\mathscr{P}(W)$ to
$\mathbb{R}$. We say that $c$ is decomposable in
U-shaped curves if, for every maximal chain $\elemset{M} \subseteq
\mathscr{P}(W)$, the restriction of $c$ to
$\elemset{M}$ is a U-shaped curve, i. e., for every $A, X, B \in
\elemset{M}$,
$A \subseteq X \subseteq B \Rightarrow \max(c(A),c(B)) \geq c(X)$.

\begin{figure}
\begin{center}
\includegraphics[width=1.0\hsize]{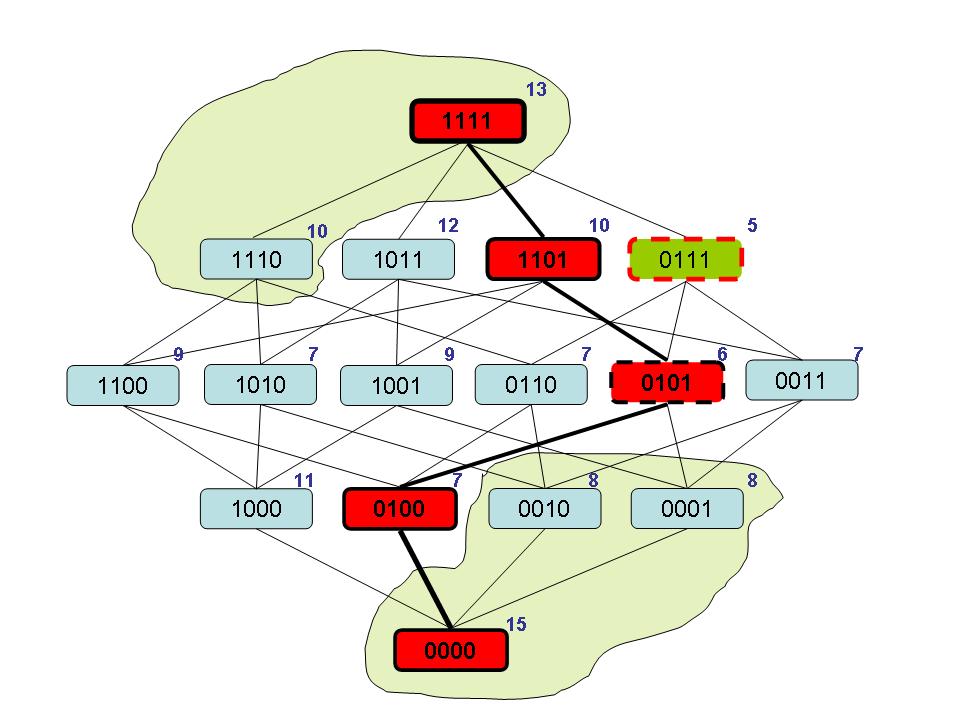}
\caption{A complete Boolean lattice $\lattice{L}$ of degree $4$ and
  the cost function decomposable
in U-shaped curves. $\UCX = \lattice{L} - \{0000, 0010, 0001, 1110,
  1111\}$ is a poset
obtained from $\lattice{L}$. A maximal chain in $\lattice{L}$ is
  emphasized. The element $0111$ is the
global minimum element and $0101$ is the local minimum element in the
  maximal chain.}
\label{fig:fig1}
\end{center}
\end{figure}

\begin{figure}
\begin{center}
\includegraphics[width=1.0\hsize]{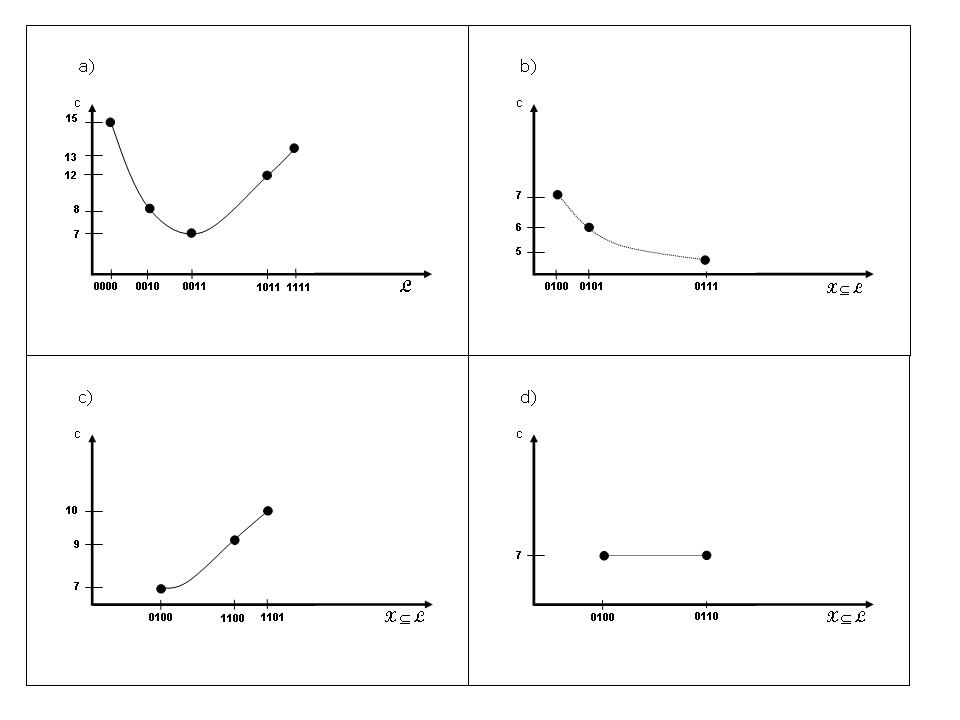}
\caption{The four possible representaion of the cost function $c$
  restricted to some maximal chains
in $\lattice{L}$ (a) and in $\UCX \subseteq \lattice{L}$ (b-d)  of
  Figure~\ref{fig:fig1}.}
\label{fig:fig2}
\end{center}
\end{figure}

Figure~\ref{fig:fig1} shows a complete Boolean lattice $\lattice{L}$
of degree $4$ with a cost function $c$ decomposable
in U-shaped curves. In this figure, it is emphasized a maximal chain
in $\lattice{L}$
and its cost function. Figure~\ref{fig:fig2} presents the curve of the
same cost function restricted
to some maximal chains in $\lattice{L}$ and in $\UCX \subseteq
\lattice{L}$.
Note the U-shape of the curves in Figure~\ref{fig:fig2}.

Our problem is to find the element (or elements) of minimum cost in a
Boolean lattice of degree $|W|$.
The full search in this space is an exponential problem, since
this space is composed by $2^{|W|}$ elements. Thus, for moderately
large $|W|$, the
full search becomes unfeasible.

%----------------------------------------
\section{The U-curve algorithm}
\noindent

The U-shaped format of the restriction of the cost function to any
maximal chain
is the key to develop a branch-and-bound algorithm, the
\newindex{U-curve algorithm}, to deal with the hard combinatorial
problem of finding subsets of minimum cost.

Let $A$ and $B$ be elements of the Boolean lattice  $\lattice{L}$.
An \newindex{interval} $[A,B]$ of $\lattice{L}$ is the subset of
$\lattice{L}$ given by
$[A,B] = \{X \in \lattice{L} : A \subseteq X \subseteq B\}$. The
elements $A$ and $B$ are called, respectively, the left and right
extremities of $[A,B]$. Intervals are very important for
characterizing decompositions in Boolean lattices \cite{Banon,Salas}.

Let $R$ be an element of $\lattice{L}$.  In this paper, intervals of
the type $[\emptyset, R]$ and
$[R,W]$ are called, respectively, lower and upper
\newindex{intervals}. The right extremity of a lower interval and
the
left extremity of an upper interval are called, respectively, lower
and upper \newindex{restrictions}.
Let $\UCRL$ and $\UCRU$ denote, respectively, collections of lower and
upper intervals.
The search space will be the \newindex{poset} $\UCX(\UCRL, \UCRU)$
obtained by eliminating the collections of lower and upper
restrictions from $\lattice{L}$, i. e.,
$\UCX(\UCRL, \UCRU) = \lattice{L} - \bigcup\{[\emptyset, R] : R \in
\UCRL\} - \bigcup\{[R, W] : R \in \UCRU\}$.
In cases in which only the lower or the upper intervals are
eliminated, the resulting search space is denoted, respectively,
by $\UCX(\UCRL)$ and $\UCX(\UCRU)$ and given, respectively, by
$\UCX(\UCRL) = \lattice{L} - \bigcup\{[\emptyset, R] : R \in \UCRL\}$
and
$\UCX(\UCRU) = \lattice{L} - \bigcup\{[R, W] : R \in \UCRU\}$.

The search space is explored by an iterative algorithm that, at each
iteration, explores a small subset
of $\UCX(\UCRL, \UCRU)$, computes a local minimum, updates the list of
minimum elements found and extends both restriction sets,
eliminating the region just explored.
The algorithm is initiated
with three empty lists:  minimum elements, lower and upper restrictions.
It is executed until the whole space is explored, i. e., until
$\UCX(\UCRL, \UCRU)$ becomes empty.
The subset of $\UCX(\UCRL, \UCRU)$ eliminated at each iteration is
defined from the
exploration of a chain,
which may be done in down-up or up-down direction.
Algorithm~\ref{alg-ucurve} describes this process. The direction
selection procedure (line~\ref{alg:alg-ucurve-direction}) can use a
random or an adaptative method. The random
method states a static probability to select the
down-up or up-down direction. The adaptative method calculates a new
probability to each direction giving more probability
to down-up direction if most of the local minima is closest to the
bottom of the lattice and up-down otherwise.

\begin{algorithm}
\caption{U-curve-algorithm()}
\label{alg-ucurve}
\begin{algorithmic}[1]
\STATE $\UCRES$ $\Leftarrow \emptyset$
\STATE $\UCRL \Leftarrow \emptyset$
\STATE $\UCRU \Leftarrow \emptyset$
\WHILE{$\UCX(\UCRL, \UCRU) \neq \emptyset$}
\STATE \emph{direction} $\Leftarrow$ Select-Direction()
\label{alg:alg-ucurve-direction}
\IF{\emph{direction} is \emph{UP}}
\STATE Down-Up-Direction($\UCRL$, $\UCRU$)
\ELSE
\STATE Up-Down-Direction($\UCRL$, $\UCRU$)
\ENDIF
\ENDWHILE
\end{algorithmic}
\end{algorithm}

\begin{figure}
\begin{center}
\includegraphics[width=1.0\hsize]{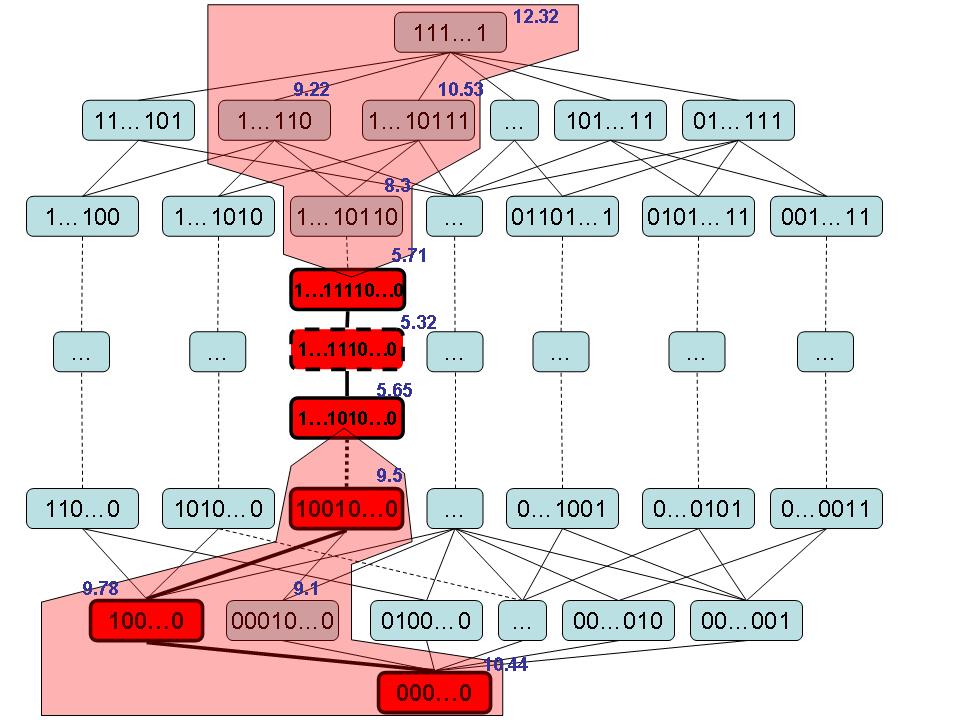}
\caption{A schematic representation of a step of the algorithm,
the detached areas represents the elements contained in a lower and
upper restrictions.}
\label{fig:fig3}
\end{center}
\end{figure}

An element $C$ of the poset $\UCX \subseteq \lattice{L}$ is called
a \newindex{minimal} element of $\UCX$, if there is no other element
$C'$ of $\UCX$ with $C' \subset C$.
In Figure~\ref{fig:fig1}, the minimal elements of $\UCX(\UCRL)$ are:
$1000$, $0100$ and $0011$.
If the down-up direction is chosen,
the Down-Up-Direction procedure is performed
(algorithm~\ref{alg-updir}):
\begin{itemize}
\item Minimal-Element procedure calculates a \newindex{minimal}
  element $B$ of the poset $\UCX(\UCRL)$.
Only the lower restriction set is used to calculate the minimal
  element $B$. An element $B$ is said to be \newindex{covered by} the
lower restriction set $\UCRL$, if $\exists R \in \UCRL : B \subseteq
  R$, and $B$ is said to be covered by the upper restriction set
$\UCRU$, if $\exists R \in \UCRU : R \subseteq B$. When the calculated
  $B$ is covered by an upper restriction, it is discarded, i.e.,
the lower restriction set is updated with $B$ and a new iteration
  begins (lines~\ref{alg:alg-updir-min1}-\ref{alg:alg-updir-min2}).
\item The down-up direction chain exploration procedure begins with a
  minimal element $B$ and flows by random selection of upper adjacent
elements from the current poset $\UCX(\UCRL,$
$\UCRU)$ until it finds the \newindex{U-curve condition},
i. e., the last element selected ($B$) has cost bigger than the
  previous one ($M$)
(lines~\ref{alg:alg-updir-chain1}-\ref{alg:alg-updir-chain2}).
\item At this point, the element $M$ is the minimum element of the chain
  explored, $A$ and $B$ are, respectively, the lower and upper
adjacent elements of $M$ (i.e., $A \subset M \subset B$ and $\{X \in
  \mathscr{P}(W): A \subset M\} = \{X \in \mathscr{P}(W): M
  \subset B = \emptyset$) by construction, $c(A) \leq c(M) \leq C(B)$.
%Note that $A$ is the empty set and $B$ is $W$, if $M$ has,
%  respectively, no lower or upper adjacent element in $\UCX(\UCRL,
%  \UCRU)$.
It can be proved that any element $C$ of $\UCX(\UCRL, \UCRU)$, with $C
  \subset A$, has cost bigger than $A$ and, any element $D$ of
$\UCX(\UCRL, \UCRU)$, with $B \subset D$, has cost bigger than $B$. By
  using this property, the lower and upper restrictions can be
  updated,
respectively, by $A$ and $B$
  (lines~\ref{alg:alg-updir-upd1}-\ref{alg:alg-updir-upd2}).
Figure~\ref{fig:fig3} shows a schematic representation of the first
  iteration of the algorithm and the elements
contained in the intervals $[\emptyset, A=1 \ldots 1010 \ldots 0]$
  and $[B=1 \ldots 11110 \ldots 0, W]$.
\item The result list can be updated with $M$
  (line~\ref{alg:alg-updir-res}) , i. e., $M$ will be included in the
  result list
if it has cost lower than the elements already saved in the list. The
  result list can save a pre-defined number of elements with
low costs or only elements with the overall minimum cost.
\item In order to prevent visiting the element $M$ more than once, a
  recursive procedure called \newindex{minimum
exhausting} procedure is performed (line~\ref{alg:alg-updir-exh})
\end{itemize}

\begin{algorithm}
\caption{Down-Up-Direction(ElementSet $\UCRL$, ElementSet $\UCRU$)}
\label{alg-updir}
\begin{algorithmic}[1]
\STATE $B \Leftarrow $ Minimal-Element($\UCRL$)
\label{alg:alg-updir-min1}
\IF{$B$ is covered by $\UCRU$}
    \STATE Update-Lower-Restriction($B$, $\UCRL$)
    \RETURN
\ENDIF    \label{alg:alg-updir-min2}
\STATE $M \Leftarrow $ \textbf{null}
\REPEAT \label{alg:alg-updir-chain1}
    \STATE $A \Leftarrow M$
    \STATE $M \Leftarrow B$
    \STATE $B \Leftarrow $ Select-Upper-Adjacent($M$, $\UCRL$,
    $\UCRU$)
\UNTIL{$c(B) > c(M)$ or $B = $ \textbf{null}}
    \label{alg:alg-updir-chain2}
\IF{$A \neq $ \textbf{null}} \label{alg:alg-updir-upd1}
    \STATE Update-Lower-Restriction($A$, $\UCRL$)
\ENDIF
\IF{$B \neq $ \textbf{null}}
    \STATE Update-Upper-Restriction($B$, $\UCRU$)
\ENDIF \label{alg:alg-updir-upd2}
\STATE Update-Results($M$) \label{alg:alg-updir-res}
\STATE Minimum-Exhausting($M$, $\UCRL$, $\UCRU$)
    \label{alg:alg-updir-exh}
\end{algorithmic}
\end{algorithm}

An element is called a \newindex{minimum exhausted} element in
$\lattice{L}$ if all its adjacents elements (upper and lower)
have cost bigger than it. This definition can be extended to the poset
$\UCX(\UCRL, \UCRU)$,
i. e., all its adjacent elements (upper and lower) in $\UCX(\UCRL,
\UCRU)$ have cost bigger than it.
In Figure~\ref{fig:fig1} we can see that
the elements $1010$, $1001$ and $0111$ are minimum exhauted elements
in $\UCX(\UCRL, \UCRU)$, but $1001$
is not a minimum exhauted element in $\lattice{L}$. In this paper, the
term minimum exhausted will be applied always refering
to a poset $\UCX(\UCRL, \UCRU)$.

\begin{algorithm}[!htb]
\caption{Minimum-Exhausting(Element $M$, ElementSet $\UCRL$,
  ElementSet $\UCRU$)}
\label{alg-minexh}
\begin{algorithmic}[1]
\STATE Push $M$ to $\UCS$
\WHILE{$\UCS$ is not empty} \label{alg:alg-minexh-loop1}
    \STATE $T \Leftarrow$ Top($\UCS$)
    \STATE \textit{MinimumExhausted} $\Leftarrow$ true
    \FORALL{$A$ adjacent of $T$ in $\UCX(\UCRL, \UCRU)$ and $A \not
      \in \UCS$} \label{alg:alg-minexh-cond2_1}
        \IF{c($A$) $\leq$ c($T$)}
            \STATE Push $A$ to $\UCS$
            \STATE \textit{MinimumExhausted} $\Leftarrow$ false
        \ELSE
            \IF{$A$ is upper adjacent of $T$}
                \STATE Update-Upper-Restriction($A$, $\UCRU$)
            \ELSE
                \STATE Update-Lower-Restriction($A$, $\UCRL$)
            \ENDIF
        \ENDIF
    \ENDFOR  \label{alg:alg-minexh-cond2_2}
    \IF{\textit{MinimumExhausted}}
        \STATE Pop $T$ from $\UCS$
        \STATE Update-Results($T$) \label{alg:alg-minexh-cond1_1}
        \STATE Update-Lower-Restriction($T$, $\UCRL$)
        \STATE Update-Upper-Restriction($T$, $\UCRU$)
    \label{alg:alg-minexh-cond1_2}
    \ENDIF \label{alg:alg-minexh-loop2}
\ENDWHILE
\RETURN
\end{algorithmic}
\end{algorithm}

The \newindex{minimum exhausting} procedure
(Algorithm~\ref{alg-minexh}) is a recursive process that visit all the
adjacent
elements of a given element $M$ and turn all of them into minimum
exhausted elements in the resulting poset $\UCX(\UCRL, \UCRU)$.
It uses a stack $\UCS$ to perform the recursive process. $\UCS$ is
initialized by pushing $M$ to it and the process is performed
while $\UCS$ is not empty
(lines~\ref{alg:alg-minexh-loop1}-\ref{alg:alg-minexh-loop2}). At each
iteration, the algorithm processes
the top element $T$ of $\UCS$: all the adjacent elements (upper and
down) of $T$ in $\UCX(\UCRL, \UCRU)$ and not in $\UCS$ are checked.
If the cost of an adjacent element $A$ is lower (or equal) than the
cost of $T$ then $A$ is pushed to $\UCS$. If the cost of $A$ is bigger
than the cost
of $T$ then one of the restriction sets can be updated with $A$, lower
restriction set if $A$ is lower adjacent of $T$ and upper restriction
set if $A$ is upper adjacent of $T$
(lines~\ref{alg:alg-minexh-cond2_1}-\ref{alg:alg-minexh-cond2_2}).
If $T$ is a minimum exhausted element in $\UCX(\UCRL, \UCRU)$, i. e.,
there is no adjacent element $A$ in $\UCX(\UCRL, \UCRU)$
with cost lower than $T$, then $T$ is removed from $\UCS$ and, also,
the
restriction sets and the result list are updated with $T$
(lines~\ref{alg:alg-minexh-cond1_1}-\ref{alg:alg-minexh-cond1_2}).
At the end of this procedure all the elements processed are
minimum-exhausted elements in $\UCX(\UCRL, \UCRU)$.

\begin{figure}
\begin{center}
\includegraphics[width=0.9\hsize]{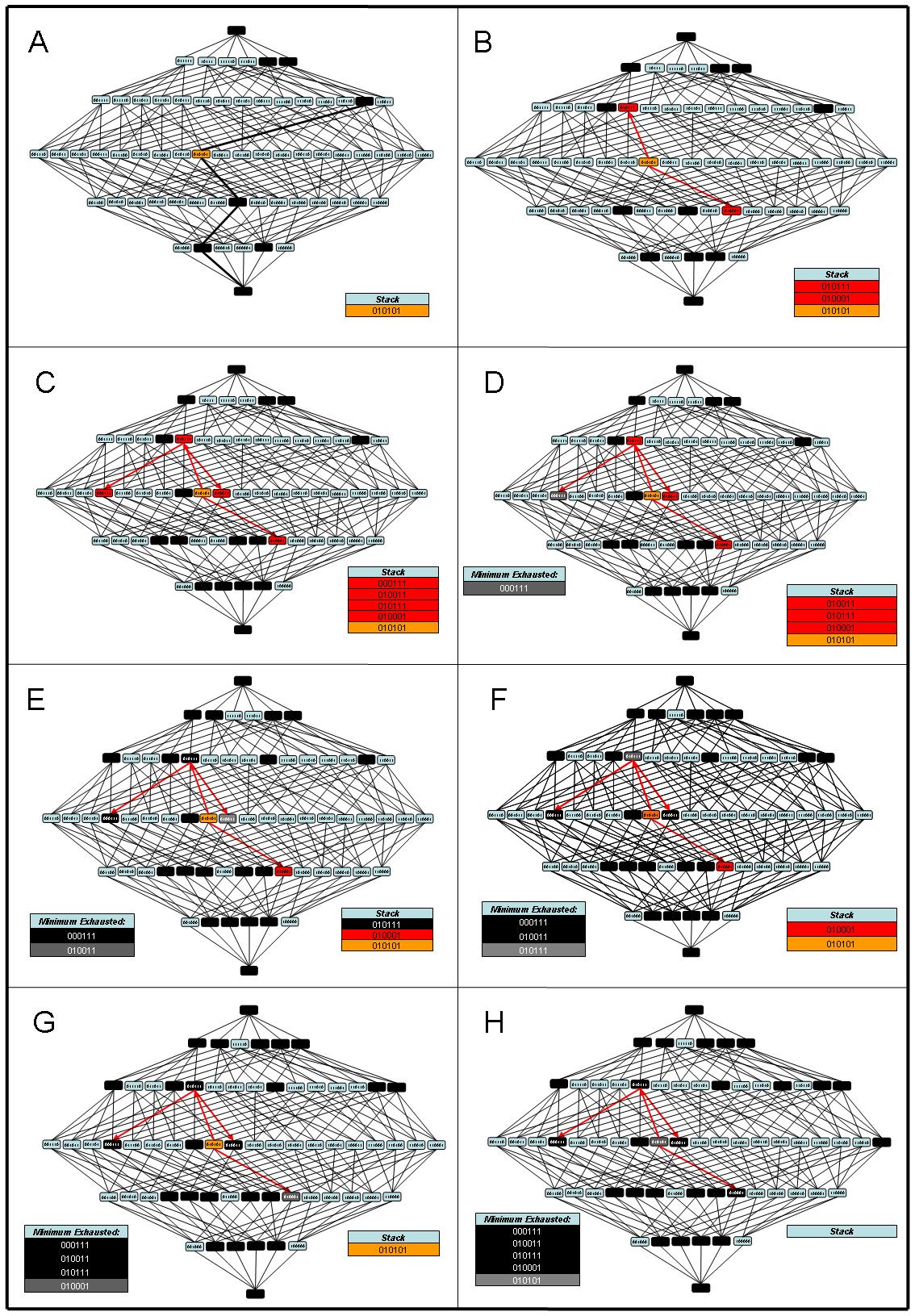}
\caption{Representation of the minimum exhausting process.}
\label{fig:fig4}
\end{center}
\end{figure}

Figure~\ref{fig:fig4} shows a graphical representation of the minimum
exhausting process.
\ref{fig:fig4}-A shows a chain construction process in up direction,
the chain has its edges emphasized.
The element $M=010101$ (orange-colored) has the minimum cost over the
chain.
The elements in black are the elements eliminated from the search
space by the restrictions obtained by the lower and upper
adjacent elements of the local minimum $M$.
The stack begins with the element $M$. Figure \ref{fig:fig4}-B shows
the first iteration of the minimum exhausting
process. The arrows in red and the elements in red indicates the
adjacents elements of $M$ (top of the stack) that
have cost lower (or equal) than it. These elements $010001$ and
$010111$ are pushed to the stack. The adjacent elements of
$M$ with cost bigger than it can update the restriction sets, i. e.,
the lower adjacent element $000101$ updates the lower
restriction set and the upper adjacent element $000101$ updates the
upper restriction set. Figure
\ref{fig:fig4}-C shows the second iteration: the adjacent elements
$010011$ and $000111$ with cost lower (or equal) than
the new top element $010111$ are pushed to the stack and the other
adjacent elements $010110$ and $011111$ with cost bigger than
$010111$ update, respectively, the lower and upper restriction sets.
In Figure \ref{fig:fig4}-D the element $000111$ is a minimum exhausted
element (grey color) in
$\UCX(\UCRL, \UCRU)$ and it is is removed from stack.
In Figure \ref{fig:fig4}-E the elements eliminated by the new
interval $[\emptyset, 000111]$ and
$[000111, W]$ are turned into black color.
At this point, $010011$ is a minimum exhausted (grey color) in
$\UCX(\UCRL, \UCRU)$ and it is removed from stack.
From Figure \ref{fig:fig4}-F to Figure \ref{fig:fig4}-H all the elements are
removed from stack and the elements removed by the
new restrictions are turned into black color. Figure \ref{fig:fig4}-H
shows all the elements removed from a single
minimum exhausted process.

The procedures to calculate minimal and maximal elements and the
procedure to update lower and upper restriction sets will be discussed
in the
next section.

%----------------------------------------
\section{Mathematical foundations}

This section introduces mathematical foundations of some modules of
the algorithm.

\subsection{Minimal and Maximal Construction Procedure}\index{Minimal
  and Maximal Construction Procedure}
Each iteration of the algorithm requires the calculation of a minimal
element in
$\UCX(\UCRL)$ or a maximal element in $\UCX(\UCRU)$. It is presented
here a simple solution for that. The next theorem
is the key for this solution.

\noindent \textbf{Theorem 1.} \textit{For every $A \in \UCX(\UCRL)$},

\begin{center}
$A \in \UCX(\UCRL) \Leftrightarrow A \cap R^c \neq \emptyset, \forall
  R \in \UCRL$.
\end{center}

\proof (in Appendix Section)

Algorithm~\ref{alg-minimal} implements the \newindex{minimal
  construction} procedure. It builds a
minimal element $C$ of the poset $\UCX(\UCRL)$. The process begins
with $C = (\underbrace{1 \ldots 1}_n)$ and $S = (\underbrace{1 \ldots
  1}_n)$ and executes a $n$-loop
(lines~\ref{alg:alg-minimal_step1}-\ref{alg:alg-minimal_step2})
trying to remove components from $C$.
At each step, a component $k, k \in \{1,\ldots,n\}$ is chosen
  exclusively from $S$ ($S$ prevents multi-selecting).
If the element $C'$ resulted from $C$ by removing the component $k$ is
  contained in $\UCX(\UCRL)$ then $C$ is updated with $C'$
(lines~\ref{alg:alg-minimal_k1}-\ref{alg:alg-minimal_k2}).

\begin{algorithm}[!htb]
\caption{Minimal-Element(ElementSet $\UCRL$)}
\label{alg-minimal}
\begin{algorithmic}[1]
\STATE $C \Leftarrow \underbrace{1 \ldots 1}_n$
\STATE $S \Leftarrow \underbrace{1 \ldots 1}_n$
\WHILE{$S \neq \underbrace{0 \ldots 0}_n$}
\label{alg:alg-minimal_step1}
    \STATE $k \Leftarrow $ random index in $\{1,\ldots, n\}$ where
    $S[k] = 1$ \label{alg:alg-minimal_rem1}
    \STATE $S[k] \Leftarrow 0$
    \STATE $C' \Leftarrow C \setminus {k}$
    \label{alg:alg-minimal_rem2}
    \STATE \textit{RemoveElement} $\Leftarrow$ true
    \label{alg:alg-minimal_k1}
    \FORALL{$R$ in $\UCRL$} \label{alg:alg-minimal_r1}
        \IF{$R^c \cap C' = \emptyset$}
            \STATE \textit{RemoveElement} $\Leftarrow$ false
        \ENDIF   
    \ENDFOR \label{alg:alg-minimal_r2}
    \IF{\textit{RemoveElement}}
        \STATE $C \Leftarrow C'$
    \ENDIF \label{alg:alg-minimal_k2}
\ENDWHILE \label{alg:alg-minimal_step2}
\RETURN $C$
\end{algorithmic}
\end{algorithm}

\noindent The minimal element calculated is
equal to $\underbrace{1 \ldots 1}_n$ when $\UCRL = \{\underbrace{1
  \ldots 1}_n\}$. At this point, the poset $\UCX(\UCRL,\UCRU)$ is
empty and the algorithm stops in the next iteration.

The next theorem proves the correctness of Algorithm~\ref{alg-minimal}
.

\noindent \textbf{Theorem 2.} \textit{The element $C$ of $\UCX(\UCRL)$
  returned by the minimal construction process (Algorithm
  \ref{alg-minimal})
is a minimal element in $\UCX(\UCRL)$.}

\proof (in Appendix Section)

The process to calculate a maximal element in $\UCX(\UCRU)$ is
\newindex{dual} to the one to calculate a minimal, i. e., it
begins with $C = \underbrace{0 \ldots 0}_n$
and, at each step, when the complement ${C'}^c$ of the resulting $C'$
has not empty interseccion
to all the elements of $\UCRU$, adds a component $k$ to $C$.

%-----------------------------------------
\subsection{Lower and Upper Restrictions Update}\index{Lower and Upper
  Restrictions Update}

The restriction sets $\UCRL$ and $\UCRU$ represent the search
space. Thus, they are updated after each new search by
the following rule: an element $A$ is added to the lower (or upper)
restriction set if all elements of
$[\emptyset, A]$ (or $[A, W]$) have costs bigger or equal to $A$.

The next theorem establishes the U-curve condition, that permits to
stop the chain construction process and to update the restriction
sets.

\noindent \textbf{Theorem 3.} \textit{Let $C_0,...,C_{k-1}, C_k$ be
  the chain constructed by Algorithm~\ref{alg-updir} (or its dual
  version).
Let $c$ be the cost function from $\lattice{L}$ to $\mathbb{R}$
  decomposable in U-shaped curves and
$c(C_k) > c(C_{k-1})$, then}

\begin{center}
$\forall{A} \in \lattice{L}, C_k \subseteq A \Rightarrow c(A) \geq
  c(C_k)$.
\end{center}

\proof (in Appendix Section)

By a similar proof to the one of Theorem 3, it can be proved that all
the elements in $\lattice{L}$
contained in $C_{k-2}$ have also cost bigger or equal to it.
Figure~\ref{fig:fig3} shows the chain obtained by the chain
construction process and the
resulted poset. The elements detached have always cost bigger than the
elements
$C_k = (1 \ldots 11110 \ldots 0)$ or $C_{k-2} = (1 \ldots 1010 \ldots
0)$.

Algorithm~\ref{alg-updlow} describes the update process of the lower
restriction set by an element $A$.
If $A$ is already covered by $\UCRL$, i. e., there exists an element
of $\UCRL$ that contains $A$ then
the process stops
(lines~\ref{alg:alg-updlow_if1_1}-\ref{alg:alg-updlow_if1_2}).
Otherwise, all the elements
in $\UCRL$ contained in $A$ are removed from $\UCRL$ and $A$ is added
to $\UCRL$
(lines~\ref{alg:alg-updlow_if2_1}-\ref{alg:alg-updlow_if2_2}). This
procedure may diminish the cardinality of the restriction set, but
does not
diminish the cardinality of the resulting poset $\UCX(\UCRL)$, since
the removed restrictions are contained in $A$.

\begin{algorithm}[!htb]
\caption{Update-Lower-Restriction(Element $A$, ElementSet $\UCRL$)}
\label{alg-updlow}
\begin{algorithmic}[1]
\IF{there exists $R$ from $\UCRL$ where $A \subseteq R$}
\label{alg:alg-updlow_if1_1}
    \RETURN
\ENDIF    \label{alg:alg-updlow_if1_2}
\FORALL{$R$ in $\UCRL$} \label{alg:alg-updlow_if2_1}
    \IF{$R \subseteq A$}
        \STATE $\UCRL = \UCRL \setminus \{R\}$
    \ENDIF
\ENDFOR
\STATE $\UCRL = \UCRL \cup \{A\}$ \label{alg:alg-updlow_if2_2}
\RETURN
\end{algorithmic}
\end{algorithm}

The upper restriction list updating procedure is \newindex{dual} to
the lower one, i. e., in this case
we look for elements contained in $A$ instead of elements that contain
$A$.

%-----------------------------------------
\subsection{Minimum Exhausting Procedure}\index{Minimum Exhausting
  Procedure}

The computation of the cost function in general is heavy.
Thus, it is desirable that each element be visited (and its cost
computed) a single time.
A way of preventing this reprocessing is to apply the minimum
exhausting procedure.
This procedure is a recursive function
(Algorithm~\ref{alg-minexh}). It uses a stack $\UCS$ to process
recursively
all the neighborhood of a given element $M$ contained in the poset
$\UCX(\UCRL, \UCRU)$. At
each recursion, it visits the upper and lower adjacent elements of
$T$, the top of $\UCS$, in $\UCX(\UCRL, \UCRU)$ and not in $\UCS$. The
adjacent elements with
cost bigger than the cost of $T$ are elements satisfying the U-curve
condition, so they can update the restriction sets and, consequently,
be removed from
the search space. The adjacent elements with cost lower or equal to
$T$ are
pushed to $\UCS$ to be processed in later iterations. Note that
elements are not
reprocessed during the exhausting procedure, since this procedure
checks if a new element explored is in an interval or in $\UCS$,
before computing its cost. If $T$ is a minimum exhausted element
in $\UCX(\UCRL, \UCRU)$ then $T$ is removed from $\UCS$.
After the whole procedure is finished, all elements processed are out
of the resulting poset $\UCX(\UCRL, \UCRU)$, so they will not be
reprocessed in the next iterations. The fact that an element can not be
reprocessed along the procedure implies that the cardinality of
$\UCX(\UCRL, \UCRU)$ is an upper limit for the
procedure number of steps.
In search spaces that are lattices with high degree, this procedure
can have to process a huge
number of elements and some heuristics should be necessary.  For
example, to stop the search
for adjacent elements smaller than a minimum after some badly
succeeded trials.

\begin{figure}
\begin{center}
\includegraphics[width=1.0\hsize]{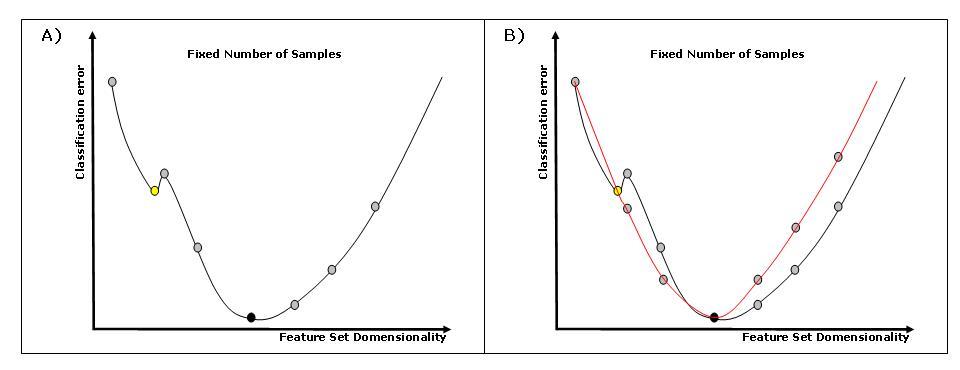}
\caption{Illustration of error curve oscillation and alternative way.}
\label{fig:fig5}
\end{center}
\end{figure}

The minimum exhausting procedure gives another interesting property to
the U-curve algorithm.
If the cost function on maximal chains are U-shaped curves with
oscillations, as illustrated in Figure~\ref{fig:fig5}-A,
the U-curve algorithm may lose a local minimum element. Note that, in
this case, the
local minimum element after the oscillation has cost smaller
than the cost of one before.
However, this minimum is not lost if there is another chain, with a
true U-shaped cost function, containing both local minimum elements.
Figure~\ref{fig:fig5}-B shows an alternative chain (chain in red) that
reaches the true minimum element of the chain (element in black).
Note that the first local minimum (element in yellow) is contained in both
chains. The true minimum, reached by the alternative chain, is obtained
exactly by the exhausting of the first minimum found.
Hence, the exhausting procedure permits to relax the class of problems
approached by the U-curve algorithm.

%----------------------------------------
\section{Experimental Results}
\noindent

In this section, some results of applications of U-curve algorithm
to feature selection are given and compared to SFFS \cite{pudil94}.
For this study several data sets were used: W-operator window design
\cite{dmartins2006},
architecture identification in genetic networks and several data sets
from the UCI Machine Learning Repository \cite{UCI}. In all cases, it was
attributed the value 3 for the parameter $\delta$ of SFFS. This
parameter is a stop criterion of SFFS. Usually, $0 < \delta \leq 3$ in
order to avoid that the algorithm stops at the first moment that it
reaches the desired dimension.
In this way, it performs more feature inclusion and
deletion before returning the subset with the desired dimension,
alleviating the nesting effect. The value $\delta = 3$ used as default
here is the same default value adopted by the original algorithm implementation
\cite{pudil94}.

All data sets used and the binary program with some documentation
can be found at the supplementary material web page
(\url{http://www.vision.ime.usp.br/~davidjr/ucurve}).

\subsection{Cost function adopted: penalized mean conditional entropy}

The Information theory was originated from Shannon´s works
\cite{Shannon} and can be employed on feature selection problems
\cite{dudahart2000chap1}. The Shannon's entropy $H$ is a measure of
randomness of a random variable $Y$ given by:

\begin{equation} \label{eq:h}
H(Y) = - \sum_{y \in Y}  P(y)  log  P(y) \textrm{,}
\end{equation}

\noindent in which $P$ is the probability distribution function and, by
convention, $0 \cdot log 0 = 0$.

The conditional entropy is given by the following equation:

\begin{equation} \label{eq:ce}
H(Y|\mathbf{X} = \mathbf{x}) = - \sum_{y \in Y} P(y|\mathbf{X} = \mathbf{x})
log P(y|\mathbf{X} = \mathbf{x})
\end{equation}

\noindent in which $\mathbf{X}$ is a feature vector and
$P(Y|\mathbf{X} = \mathbf{x})$ is the conditional
probability of $Y$ given the observation of an instance $\mathbf{x}
\in \mathbf{X}$. Finally, the mean conditional entropy of $Y$
given all the possible instances $\mathbf{x} \in \mathbf{X}$ is given
by:

\begin{equation} \label{eq:mce}
E[H(Y|\mathbf{X})] = \sum_{\mathbf{x} \in \mathbf{X}} P(\mathbf{x})
H(Y|\mathbf{x})
\end{equation}

Lower values of $H$ yield better feature subspaces (i.e., the lower $H$,
the larger is the information gained about $Y$ by observing $\mathbf{X}$).

In practice, $H(Y)$ and $H(Y|\mathbf{X})$ are estimated. A way to
embed the error estimation, committed by using feature vectors
with large dimensions and insufficient number of samples, is to
atribute a high entropy (i.e., penalize) to the rarely observed
instances. The penalization adopted here consists
in changing the conditional probability distribution of the instances
that present just a unique observation to uniform distribution
(i.e., the highest entropy). This makes sense because if an instance
$\mathbf{x}$ has only 1 observation, the value of $Y$ is fully determined
(i.e., $H(Y|\mathbf{X} = \mathbf{x}) = 0$), but
the confidence
about the real distribution of $P(Y|\mathbf{X} = \mathbf{x})$ is very low.
Adopting this penalization, the estimation of the mean
conditional entropy becomes:

\begin{equation} \label{eq:mce2}
\hat{E}[H(Y|\mathbf{X})] =
\frac{N}{t} +
\sum_{\mathbf{x} \in
\mathbf{X}:\hat{P}(\mathbf{x}) > \frac{1}{t}}
\hat{P}(\mathbf{x}) \hat{H}(Y|\mathbf{x})\textrm{,}
\end{equation}

\noindent in which $t$ is the number of training samples and $N$ is the
number of instances with $P(\mathbf{x}) = \frac{1}{t}$
(i.e., just one observation). In this formula, it is assumed that the logarithm
base
is the number of possible classes $|Y|$, thus, normalizing the entropy
values to the interval $[0, 1]$.
This cost function exhibits U-shaped curves, since,
for a sufficiently large dimension, the number of instances with a single
observation starts to increase, increasing the penalization and,
consequently, increasing the cost
function value (i.e., next features included do not give enough
information to compensate the error estimation).

\subsection{Data sets description}

\subsubsection{W-operator window design}

the W-operator window design problem
consists in looking
for subsets of a size $n$ window for which the designed operator has the
lowest estimation
error (i. e., the transformed images generated by the operator are as
similar as possible of the expected images). The training samples were
obtained from the images presented in \cite{dmartins2006}. It is composed
by 20 files with 18,432 samples each. There are 16 features assuming
binary values and two
classes.

\subsubsection{Biological classification}

the biological classification problem
studied is the problem
of estimating a subset of predictor genes for a specific target gene
from a time-course
microarray experiment. The data set used for the tests is
the one presented in paper \cite{estrogen_edliu04}. They are normalized
and quantized in $3$ levels using the same method described in
\cite{barreranando}. The subset of predictors is obtained from a
set of $27$ genes. Thus, there are 27 features assuming three distinct
values and three possible classes.
It is composed by 10 files with 15 samples each.

\subsubsection{UCI Machine Learning Repository}

UCI Machine Learning Repository data sets considered are:
\textit{pendigits, votes, ionosphere, dorothea\_filtered, dexter\_filtered,
spambase, sonar and madelon}. For
all data sets, the feature values were normalized by subtracting them
from their respective means
and dividing them by their respective standard deviations. After that,
all values were binarized (i.e., associated to 0,
if the normalized value is non-positive, and to 1, otherwise). Except
for dorothea\_filtered and dexter\_filtered, all features
were taken into account. The
\textit{dorothea\_filtered} and \textit{dexter\_filtered} are files
post-processed from \textit{dorothea} and \textit{dexter} data sets,
respectively. In the \textit{dorothea} and \textit{dexter} data sets,
most features display null value for almost every sample. So,
\textit{dorothea\_filtered} considered only the features with 100 or
more non-null values, while \textit{dexter\_filtered} considered the
features with 50 or more non-null values.

A description of each data set is presented in the following list:

\begin{itemize}

\item {\em pendigits:} composed by 7494 samples, 16 binary features
  and 10 classes;

\item {\em votes:} composed by 435 samples, 16 ternary features and 2
  classes;

\item {\em ionosphere:} composed by 351 samples, 34 binary features
  and 2 classes;

\item {\em dorothea\_filtered:} composed by 800 samples, 38 binary
  features and 2 classes;

\item {\em dexter\_filtered:} composed by 300 samples, 48 binary features and
  2 classes;

\item {\em spambase:} composed by 4601 samples, 57 binary features and
  2 classes;

\item {\em sonar:} composed by 208 samples, 60 binary features and 2
  classes;

\item {\em madelon:} composed by 2000 samples, 500 binary features and
  2 classes.

\end{itemize}

\subsection{Results}

The feature selection problem may have cost functions with chains that
present oscillations
and there is no theoretical guaranty of the existence of alternative
chains to achieve the local
minima lost because of the oscillations. However, these cases were
tested experimentally
and in all observed cases the  minimum exhausting procedure could find
the local minimum elements using alternative chains.
We have examined $100,000$ random curves in all data sets studied. For
example, in the W-operator
window design almost $24,000$ curves ($24\%$) contains oscillatory parts and
in the biological classifier design almost $15,000$ curves ($15\%$) contain
oscillatory parts. For
all these oscillatory curves and also for those found in
the UCI data sets, the minimum exhausting procedure got the local minimum
by alternative chains.

The results of the U-curve algorithm are divided in two sets: i - until
it beats the SFFS result (UC); ii- until the search space is
completely processed (UCC). The U-curve algorithm is stochastic and
at each test it can reach the best result in different processing
time. So, the U-curve was processed $5$ times for each test and the
quantitative results presented are means of values gotten in these $5$
processes. The machine used for the tests was an AMD Turion 64 with 2Gb of RAM.

In the following, each of the three experiments performed is summarized
by a table and all
these tables have the same structure. The first column presents the
winner of the comparison
of SFFS with UC. The other columns present the cost in terms of
processed nodes and computational time of SFFS, UC and UCC.

Table~\ref{tab:uc_woperator} shows the results for the W-operator
window design experiment. Twenty tests were performed using the
available training samples.
UC beats SFFS in 8 of the 20 tests and reaches the same result in the
remaining ones. In these last cases, both reach the global minimum
element. In all cases,
UC processes a smaller number of nodes, in a smaller time, than SFFS.
The complete search (UCC) frequently needs to process more nodes
($17/20$), taking more time ($19/20$),
than SFFS.

\begin{table}[htp]
\caption{Comparison between SFFS and U-curve results for the
  W-operator window design.}
\label{tab:uc_woperator}
\centering
\begin{tabular}{|c|c|c|c|c|c|c|c|}
\hline Test & Winner & \multicolumn{3}{c|}{Computed nodes} &
\multicolumn{3}{c|}{Time(sec.)} \\
\hline      &        & SFFS & UC & UCC & SFFS & UC & UCC \\
\hline 1 & EQUAL & $358$ & $73$ & $373$ & $8$ & $2$ & $393$ \\
\hline 2 & EQUAL & $333$ & $31$ & $154$ & $7$ & $1$ & $392$ \\
\hline 3 & EQUAL & $417$ & $17$ & $137$ & $10$ & $1$ & $393$ \\
\hline 4 & UC    & $435$ & $58$ & $5,965$ & $9$ & $1$ & $541$ \\
\hline 5 & UC    & $357$ & $101$ & $223$ & $7$ & $3$ & $385$ \\
\hline 6 & UC    & $384$ & $66$ & $345$ & $9$ & $2$ & $399$ \\
\hline 7 & EQUAL & $302$ & $111$ & $266$ & $6$ & $4$ & $392$ \\
\hline 8 & UC    & $1,217$ & $158$ & $13,963$ & $21$ & $2$ & $591$ \\
\hline 9 & UC    & $330$ & $31$ & $274$ & $8$ & $1$ & $385$ \\
\hline 10 & EQUAL & $406$ & $113$ & $825$ & $10$ & $4$ & $408$ \\
\hline 11 & EQUAL & $329$ & $70$ & $544$ & $7$ & $2$ & $387$ \\
\hline 12 & EQUAL & $336$ & $17$ & $17$ & $8$ & $0.5$ & $0.5$ \\
\hline 13 & EQUAL & $310$ & $26$ & $26$ & $8$ & $1$ & $384$ \\
\hline 14 & UC    & $328$ & $67$ & $67$ & $8$ & $4$ & $421$ \\

\hline 15 & EQUAL & $425$ & $66$ & $671$ & $8$ & $1$ & $391$ \\
\hline 16 & UC    & $333$ & $31$ & $151$ & $8$ & $1$ & $377$ \\
\hline 17 & EQUAL & $1,257$ & $659$ & $11,253$ & $31$ & $16$ & $717$
\\
\hline 18 & UC    & $336$ & $39$ & $218$ & $7$ & $1$ & $385$ \\
\hline 19 & EQUAL & $296$ & $32$ & $137$ & $6$ & $2$ & $379$ \\
\hline 20 & EQUAL & $323$ & $31$ & $151$ & $8$ & $2$ & $376$ \\
\hline
\end{tabular}
\end{table}

Table~\ref{tab:uc_microarray} shows the results for the biological
classifier design experiment. Ten tests were performed
using different target genes. In these examples, the complete search
space is quite big ($2^{27}$ nodes).
SFFS reaches the best element, equalling UC, only $3/10$ times. The
processing of the whole space (UCC)
improved the result of UC in $7/10$ times. UC processed many more nodes
than SFFS, but their computational
times are very similar. This happens because these experiments involve
small number of samples and, therefore, the computational time spent
to process a node is very small. The pre-processing overhead is the major
responsible for the time consuming in this case.

\begin{table}[htp]
\caption{Comparison between SFFS and U-curve results
for the biological classification design.}
\label{tab:uc_microarray}
\centering
\begin{tabular}{|c|c|c|c|c|c|c|c|}
\hline Test & Winner & \multicolumn{3}{c|}{Computed nodes} &
\multicolumn{3}{c|}{Time(sec.)} \\
\hline      &        & SFFS & UC & UCC & SFFS & UC & UCC \\
\hline  1 & EQUAL & $135$ & $777$ & $9,964$ & $0.5$ & $0.6$ & $3.1$ \\
\hline  2 & UC    & $135$ & $9252$ & $30,724$ & $0.5$ & $2.1$ &
$11.2$\\
\hline  3 & UC    & $135$ & $1037$ & $9,410$ & $0.5$ & $0.6$ & $3.1$\\
\hline  4 & UC    & $164$ & $786$ & $9,276$ & $0.5$ & $0.6$ & $3.1$\\
\hline  5 & UC    & $281$ & $247$ & $6,126$ & $0.5$ & $0.6$ & $1.5$\\
\hline  6 & EQUAL & $135$ & $2675$ & $11,031$ & $0.5$ & $0.7$ &
$7.3$\\
\hline  7 & EQUAL & $135$ & $998$ & $10,836$ & $0.5$ & $0.6$ & $6.9$\\
\hline  8 & UC    & $135$ & $463$ & $5,381$ & $0.5$ & $0.5$ & $1.5$ \\
\hline  9 & UC    & $135$ & $246$ & $4,226$ & $0.5$ & $0.5$ & $1.5$ \\
\hline  10 & UC    & $191$ & $474$ & $8,930$ & $0.5$ & $0.5$ & $2.9$
\\
\hline
\end{tabular}
\end{table}

Table~\ref{tab:uc_public} shows the results of 8 tests using public
datasets. For each test, the value in parenthesis
is the number of features (n) in the data set. For tests with high
number of features, the results
for the complete search (UCC) are not available. We can see that UC
obtained better results
than SFFS in $6/8$ of the tests and equal results in two tests with small
number of features. In these two cases, SFFS reaches the best result but
UC reaches them faster, processing less nodes.

\begin{table}[htp]
\caption{Comparison between SFFS results and U-curve algorithm 
for the UCI Machine Learning Repository data sets.}
\label{tab:uc_public}
\centering
\begin{tabular}{|c|c|c|c|c|c|c|c|}
\hline Test & Winner & \multicolumn{3}{c|}{Computed nodes} & \multicolumn{3}{c|}{Time(sec.)} \\
\hline      &        & SFFS & UC & UCC & SFFS & UC & UCC \\
\hline pendigits (16) & EQUAL & $358$ & $124$ & $1,292$ & $5$ & $1$ & $19$ \\ 
\hline votes (16) & EQUAL & $128$ & $81$ & $12,670$ & $0.03$ & $0.02$ & $87$ \\ 
\hline ionosphere (34) & UC & $4,782$ & $1,139$ & NA & $1$ & $0.25$ & NA \\ 
\hline dorothea\_filtered (37) & UC & $7,004$ & $799$ & NA & $10$ & $1$ & NA \\ 
\hline dexter\_filtered (48) & UC & $8,071$ & $596$ & NA & $3$ & $1$ & NA \\ 
\hline spambase (57) & UC & $24,265$ & $1,608$ & NA & $441$ & $21$ & NA \\ 
\hline sonar (60) & UC & $540$ & $784$ & NA & $0.09$ & $0.10$ & NA \\ 
\hline madelon (500) & UC & $66,403$ & $159,745$ & NA & $1,000$ & $3,008$ & NA \\ 
\hline
\end{tabular}
\end{table}

These results show that UC is more efficient than SFFS for
low order problems, obtaining the same results with less processing. For
high order problems,
UC is more accurate, but in some cases it process more nodes and takes
more time.

%----------------------------------------
\section{Conclusion}
\noindent

This paper introduces a new combinatorial problem, the Boolean
U-curve optimization problem, and presents a stochastic
branch-and-bound solution for it,
the U-curve algorithm. This algorithm gives the optimal elements of a
cost function
decomposable in U-shaped chains, that may even be oscillatory in a
given sense.
This model permits to describe the feature
selection problem in the context of pattern recognition. Thus, the
U-curve algorithm
constitutes a new tool to approach feature selection problems.

The U-curve algorithm explores the domain and cost function particular
structures. The Boolean nature
of the domain permits to represent
the search space by a collection of upper and lower restrictions. At
each iteration, a beginning of
chain node is computed from the search space restrictions. The current
explored chain is
constructed from this node by choosing upper or lower adjacent
nodes. The choice of a beginning of
chain and of an adjacent node usually has several options and one of
them is taken randomly.
The cost function and domain structure permit to make cuts in the
search space, when a local
minimum is found in a chain. After a local minimum is found, all local
minimum nodes connected to it
are computed, by the minimum exhausting procedure, and the
corresponding cuts, by up-down intervals, executed.
The adjacency and connectivity relations adopted are the ones of the
search space Hesse diagram,
that is a graph in which the connectivity is induced by the partial
order relation.
The minimum exhausting procedure avoids that a node be visited more
than once and generalizes the
algorithm to cost functions decomposable in some class of U-shaped
oscillatory chain functions.
The procedures of the U-curve algorithm are supported by formal
results.

In fact, the U-curve optimization technique constitutes a new
framework to study a family of optimization problems. The restrictions
representation and the intervals cut, based on Boolean lattice
properties,
constitutes a new optimization structure for combinatorial problems,
with
properties not found in conventional tree representations.

The U-curve was applied to practical problems and compared to
SFFS. The experiments involved window operator design, genetic network
identification and six public data sets obtained from the UCI repository.
In all experiments, the results of the U-curve algorithm were equal or
better than those obtained from SFFS in precision and, in many cases,
even in performance. The results of the U-curve algorithm considered for
comparison are the mean of
several executions for the same input data, since it is a stochastic
algorithm that may
have different performances at each run.

The efficiency of the U-curve algorithm depends on the relative
position of the local
minima on the search space. The algorithm is more efficient when the
local minima are near
the search space extremities. The worst cases are the ones in which
the local minima are near
the middle of the lattice.

The results obtained until now are encouraging, but the present
version of the U-curve algorithm
is not a fast solution for high dimension problems with many local
minima in the center of the
search space lattice. The efficient addressing of these problems in
the U-curve optimization
approach opens a number of subjects for future researches such as: to
develop additional cuts to the
branch-and-bound formulation; to design and estimate distributions for
the random parameters
used in the choice of beginning nodes or adjacent paths in the
construction of a chain, with
the goal of reaching earlier to the best nodes; to build parallelized
versions of the algorithm; and others.

\section*{Appendix}

\noindent \textbf{Theorem 1.} \textit{For every $A \in \UCX(\UCRL)$},

\begin{center}
$A \in \UCX(\UCRL) \Leftrightarrow A \cap R^c \neq \emptyset, \forall
  R \in \UCRL$.
\end{center}

\proof
\begin{eqnarray*}
A \in \UCX(\UCRL) & \Leftrightarrow &  A \in \lattice{L} -
\bigcup\{[\emptyset, R] : R \in \UCRL\} \nonumber\\
                                 &\Leftrightarrow &  A \notin
\bigcup\{[\emptyset, R] : R \in \UCRL\} \nonumber\\
                                 &\Leftrightarrow &  A \notin
     [\emptyset, R], \forall R \in \UCRL \nonumber\\
                                 &\Leftrightarrow &  A \not \subseteq
     R, \forall R \in \UCRL \nonumber\\
                                 &\Leftrightarrow &  A \cap R^c \neq
     \emptyset, \forall R \in \UCRL \nonumber\\
\end{eqnarray*}
\QEDopen

\noindent \textbf{Theorem 2.} \textit{The element $C$ of $\UCX(\UCRL)$
  returned by the minimal construction process (Algorithm
  \ref{alg-minimal})
is a minimal element in $\UCX(\UCRL)$}

\proof
By looking into the steps of the minimal construction procedure:
\begin{itemize}
\item Lines ~\ref{alg:alg-minimal_k1}-\ref{alg:alg-minimal_k2}
  guarantee that at any step of the procedure the
resulted $C$ is contained in $\UCX(\UCRL)$, i. e., it is updated only
  when the resulted $C'$ satisfies the condition
shown in Theorem 1.
\item Let $C_1, \ldots, C_n$ be the sequence of resulting elements at
  each step $i$ ($i=1,\ldots, n$) and
$C_0=\underbrace{1 \ldots 1}_n$ be the initial element.
As an index $k$ is chosen to be removed from
  $C_{i-1}$(lines~\ref{alg:alg-minimal_rem1}-\ref{alg:alg-minimal_rem2})
at each step $i$, it implies that
$C_n \subseteq C_{n-1} \subseteq \ldots \subseteq C_0$.
\item Proving that the resulting element $C_n$ is mimimal in
$\UCX(\UCRL)$ is equivalent of proving that $\forall l \in C_n, C_n
  \setminus \{l\} \not \in \UCX(\UCRL)$.
\item Let $k=l, l \in C_n$ and $i$ be the step of the procedure
when the index $l$ is chosen to be removed from $C_{i-1}$. $C_n
\subseteq C_i$ and $l \in C_n$ imply that
$l \in C_i$, i. e., $l$ cannot be removed from $C_{i-1}$ at the end of
step $i$. This is avoided by
the algorithm
(lines~\ref{alg:alg-minimal_r1}-\ref{alg:alg-minimal_r2}), when there
exists an element
$R \in \UCRL$ with $R^c  \cap (C_{i-1} \setminus \{l\}) = \emptyset$.
As $C_n \setminus \{l\} \subseteq C_{i-1} \setminus \{l\}$, then
$R^c \cap (C_n \setminus \{l\}) = \emptyset$ and, by Theorem 1,
$C_n \setminus \{l\} \not \in \UCX(\UCRL)$. This implies that $C_n$ is
a minimal element in $\UCX(\UCRL)$.
\end{itemize}
\QEDopen

\noindent \textbf{Theorem 3.} \textit{Let $C_0,...,C_{k-1}, C_k$ be
  the chain constructed by Algorithm~\ref{alg-updir} (or its dual
  version).
Let $c$ be the cost function from $\lattice{L}$ to $\mathbb{R}$
  decomposable in U-shaped curves and
$c(C_k) > c(C_{k-1})$. It is true that,}

\begin{center}
$\forall{A} \in \lattice{L}, C_k \subseteq A \Rightarrow c(A) \geq
  c(C_k)$.
\end{center}

\proof
Suppose that $\exists B \in \lattice{L}, C_k \subseteq B$ and $c(B) <
c(C_k)$. It contradicts the hypothesis that
$c$ is a function decomposable in U-shaped curves, since $C_{k-1}
\subseteq C_k \subseteq B$, but
$\max(c(C_{k-1}),c(B))$ is either $c(C_{k-1}) < c(C_k)$ or $c(B) <
c(C_k)$, contradicting
$\max(c(C_{k-1}),c(B)) > c(C_k)$.
\QEDopen

%----------------------------------------
\section*{Acknowledgement}
\noindent
The authors are grateful to FAPESP (99/12765-2, 01/09401-0, 04/03967-0
and 05/00587-5),
CNPq (300722/98-2, 468 413/00-6, 521097/01-0 474596/04-4 and
491323/05-0) and CAPES for financial support.
This work was partially supported by grant 1 D43 TW07015-01 from the
National Institutes of Health, USA. We also thank
Helena Brentani by her helpful in the data for biological analysis and
Roberto M. Cesar Jr. by his helpful in SFFS comparisons. The data sets
used to generate the Table~\ref{tab:uc_public} results were obtained from
UCI Machine Learning Repository \cite{UCI}.

\bibliographystyle{plain}
\bibliography{mrbib}

\begin{biography}[{\includegraphics[width=1in,height=1.25in,clip,keepaspectratio]{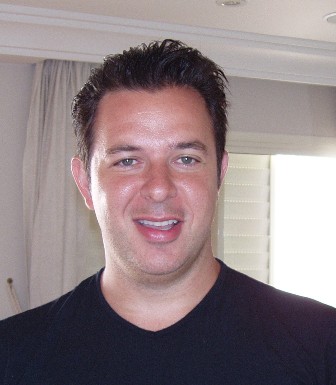}}]{Marcelo
    Ris}
received a B.Sc. in Computer Science
(Universidade de S\~{a}o Paulo - USP, Brazil) and a M.Sc. in Computer
Science
(Universidade de S\~{a}o Paulo - USP, Brazil). His main research
topics are in
bioinformatics, including algorithms design for gene network
identification,
pattern recognition for computer vision and algorithm parallelism. He
is
currently a Ph.D student on Bioinformatics at Universidade de S\~{a}o
Paulo -
USP, Brazil and Hospital do C\^{ancer} - Brazill.
\end{biography}

% if you will not have a photo
\begin{biography}[{\includegraphics[width=1in,height=1.25in,clip,keepaspectratio]{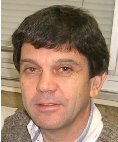}}]{Junior
    Barrera}
received a B.Sc. in Electrical Engineering (Universidade de
S\~{a}o Paulo - USP, Brazil), a M.Sc in Applied Computing (Instituto
Nacional
de Pesquisas Espaciais - INPE, Brazil) and a Ph.D. in Electrical
Engineering (Universidade de S\~{a}o Paulo - USP, Brazil). His main
research
topics are study of lattice operator representation and design,
lattice
dynamical systems, image processing, bioinformatics and computational
biology. He is currently a full professor at the Department of
Computer
Science of IME - USP and president of the Brazilian Society for
Bioinformatics and Computational Biology.
\end{biography}

% insert where needed to balance the two columns on the last page
%\newpage

\begin{biography}[{\includegraphics[width=1in,height=1.25in,clip,keepaspectratio]{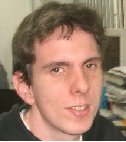}}]{David
    C. Martins Jr}
received a B.Sc. in Computer Science
(Universidade de S\~{a}o Paulo - USP, Brazil) and a M.Sc. in Computer
Science
(Universidade de S\~{a}o Paulo - USP, Brazil). His main research
topics are in
pattern recognition for computer vision and bioinformatics, including
but
not limited to gene network identification. He is currently a Ph.D
student
on Computer Science at IME - USP and recently he did a research stage
at the Genomics Signal Processing Laboratory - Texas A. \&
M. University
during one year.
\end{biography}

\end{document}